\documentclass{sig-alternate-05-2015}

\usepackage{tikz}
\usepackage{pgfplots}
\usepackage{balance}
\usetikzlibrary{shapes,arrows}

\newcommand{\sentence}[1]{\emph{#1}}

\def\sharedaffiliation{
\end{tabular}
\begin{tabular}{c}}

\makeatletter
\renewcommand*{\@fnsymbol}[1]{\ensuremath{\ifcase#1\or \bigstar\or \dagger\or
    \ddagger\or
    \mathsection\or \mathparagraph\or \|\or **\or \dagger\dagger
\or \ddagger\ddagger \else\@ctrerr\fi}}
\makeatother

\begin{document}

\CopyrightYear{2016} 
\setcopyright{rightsretained} 
\conferenceinfo{KDD '16}{August 13-17, 2016, San Francisco, CA, USA} 
\isbn{978-1-4503-4232-2/16/08}
\doi{http://dx.doi.org/10.1145/2939672.2939801}

\clubpenalty=10000 
\widowpenalty = 10000

\title{Smart Reply: Automated Response Suggestion for Email}

\numberofauthors{6} 
\author{
\alignauthor
Anjuli Kannan\thanks{Equal contribution.} \\
\alignauthor
Karol Kurach\footnotemark[1] \\
\alignauthor
Sujith Ravi\footnotemark[1] \\ \and
\alignauthor
Tobias Kaufmann\footnotemark[1] \\
\alignauthor
Andrew Tomkins \\
\alignauthor
Balint Miklos \\ \and
\alignauthor
Greg Corrado \\
\alignauthor
L\'aszl\'o Luk\'acs \\
\alignauthor
Marina Ganea \\ \and
\alignauthor
Peter Young \\
\alignauthor
Vivek Ramavajjala \\
\and
\sharedaffiliation{Google} \\
\email{\{anjuli, kkurach, sravi, snufkin\}@google.com}\\
} 

\date{12 Feb 2016}

\maketitle
\begin{abstract}

In this paper we propose and investigate a novel end-to-end method for
automatically generating short email responses, called Smart Reply. It
generates semantically diverse suggestions that can be used as complete email
responses with just one tap on mobile.  The system is currently used in
\emph{Inbox by Gmail} and is responsible for assisting with $10$\% of all mobile
responses.  It is designed to work at very high throughput and process hundreds
of millions of messages daily. The system exploits state-of-the-art,
large-scale deep learning.

We describe the architecture of the system as well as the challenges that we
faced while building it, like response diversity and scalability.
We also introduce a new method for semantic clustering of user-generated
content that requires only a modest amount of explicitly labeled data.

\end{abstract}

\printccsdesc

\keywords{Email; LSTM; Deep Learning; Clustering; Semantics}

\section{Introduction}
\label{sec:intro}

Email is one of the most popular modes of communication on the Web. Despite the
recent increase in usage of social networks, email continues to be the primary
medium for billions of users across the world to connect and share information
\cite{ipsos}. With the rapid increase in email overload, it has become
increasingly challenging for users to process and respond to incoming messages.
It can be especially time-consuming to type email replies on a mobile device.

An initial study covering several million email-reply pairs showed that
$\sim$25\% of replies have 20 tokens or less.  Thus
we raised the following question: can we assist users with composing these short
messages? More specifically, would it be possible to suggest brief responses
when appropriate, just one tap away?

\begin{figure}[h]
\centering
\includegraphics[scale=0.11]{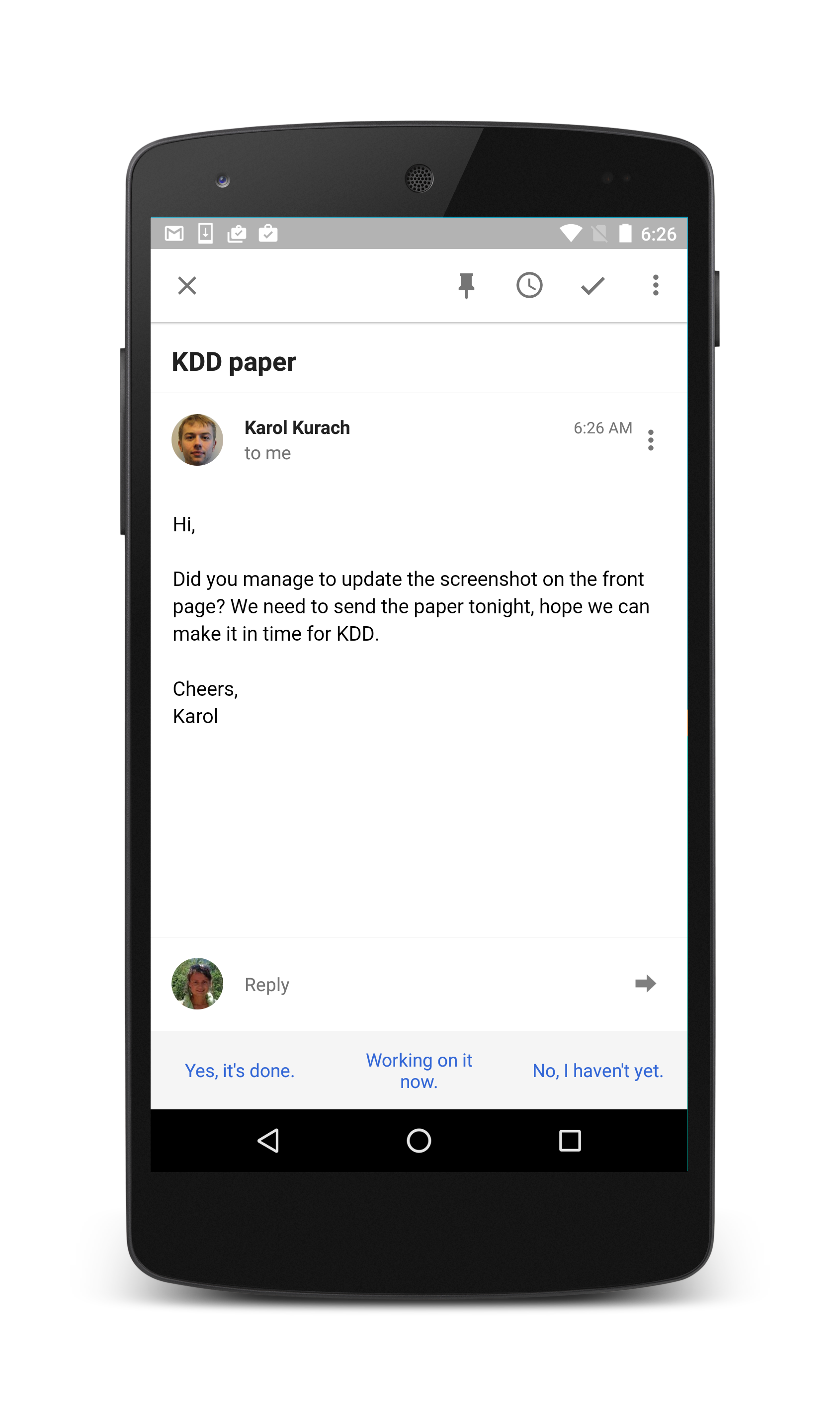}
\caption{Example Smart Reply suggestions.}
\label{fig:screenshot}
\end{figure}

\newpage 

To address this problem, we leverage the sequence-to-sequence learning framework
\cite{SutskeverVL14}, which uses long short-term memory networks (LSTMs)
\cite{lstm} to predict sequences of text.  Consistent with the approach of the
Neural Conversation Model \cite{vinyals_conversation}, our {\it input} sequence
is an incoming message and our {\it output} distribution is over the
space of possible replies.  Training this framework on a large corpus of
conversation data produces a fully generative model that can produce a response
to any sequence of input text.  As \cite{vinyals_conversation} demonstrated on
a tech support chat corpus, this distribution can be used to decode coherent,
plausible responses.

However, in order to deploy such a model into a product used globally by
millions, we faced several challenges not considered in previous work:

\begin{itemize}
\item{{\bf Response quality} How to ensure that the individual response options are
{\it always} high quality in language and content.}
\item{{\bf Utility} How to select multiple options to show a user so as
to maximize the likelihood that one is chosen.}
\item{{\bf Scalability} How to efficiently process millions of messages per day
while remaining within the latency requirements of an email delivery system.}
\item{{\bf Privacy} How to develop this system without ever inspecting the data
except aggregate statistics.}
\end{itemize}

\tikzstyle{decision} = [diamond, draw, 
    text width=4.5em, text badly centered, node distance=2.5cm, inner sep=0pt]
\tikzstyle{block} = [rectangle, draw, 
    text width=5em, text centered, rounded corners, minimum height=4em]
\tikzstyle{line} = [draw, -latex']
\tikzstyle{cloud} = [draw, ellipse,fill=red!20, node distance=3cm,
    minimum height=2em]
\tikzstyle{datasource} = [draw, black, thick, cylinder, alias=cyl , shape border rotate=90, aspect=0.3, minimum width=5em] 

\begin{figure}
\begin{tikzpicture}[node distance = 2cm, auto]
	\node [rectangle] (init) {\includegraphics[width=.07\textwidth]{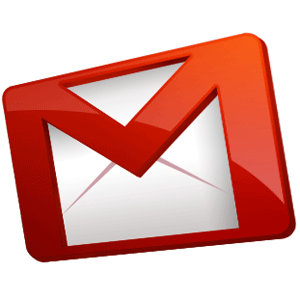}};
	\node [block, below of=init] (preprocess) {Preprocess Email};
    \node [decision, below of=preprocess] (triggering) {Trigger response?};
    \node [block, left of=triggering, node distance=3cm, fill=red!20] (noreply) {No Smart Reply};
    \node [block, below of=triggering, node distance=2.5cm] (lstm) {Response
    selection (LSTM)};
    \node [datasource, right of=lstm, align=center, node distance=3cm] (responselist) {Permitted \\ responses \\ and clusters};
	\node [block, below of=lstm] (diversity) {Diversity selection};	
	\node [block, left of=diversity, node distance=3cm, fill=green!20] (success) {Smart Reply Suggested};

    \path [line] (init) -- (preprocess);
    \path [line] (preprocess) -- (triggering);
    \path [line] (triggering) -- node [near start] {no} (noreply);
	\path [line] (triggering) -- node [near start] {yes} (lstm);
	\path [line] (lstm) -- (diversity);
  \path [line] (diversity) -- (success);
    \path [line,dashed] (responselist) -- (lstm);
	\path [line,dashed] (responselist) |- (diversity);
\end{tikzpicture}
\caption{Life of a message. The figure presents the overview of inference.} \label{fig:inference_overview}
\end{figure}
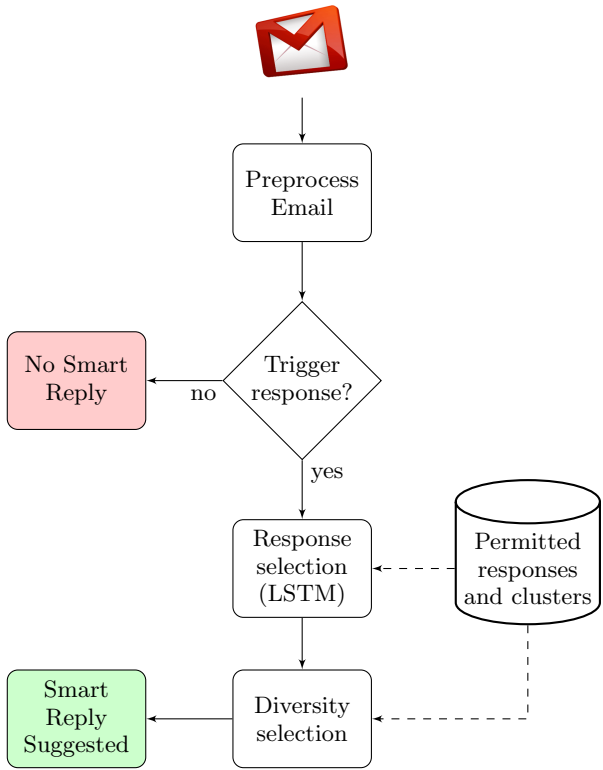

To tackle these challenges, we propose Smart Reply (Figure ~\ref{fig:screenshot}),
a novel method and system for automated email response suggestion.  Smart Reply
consists of the following components, which are also shown in
Figure~\ref{fig:inference_overview}:

\begin{enumerate}

\item {\bf  Response selection:} At the core of our system, an LSTM neural
  network processes an incoming message, then uses it to predict the most likely
  responses.  LSTM computation can be expensive, so we improve {\it scalability}
  by finding only the approximate best responses.  We explain the model
  in detail in Section \ref{sec:response_selection}.

\item {\bf  Response set generation:} To deliver high {\it response quality}, we
  only select responses from response space which is generated offline using a semi-supervised graph
  learning approach. This is discussed in Section \ref{sec:response_generation}.

\item {\bf Diversity:} After finding a set of most likely responses from the
  LSTM, we would like to choose a small set to show to the user that maximize
  the total {\it utility}.  We found that enforcing diverse semantic intents
  is critical to making the suggestions useful. Our method for
  this is described further in Section \ref{sec:diversity}.

\item {\bf Triggering model:} A feedforward neural network decides whether
  or not to suggest responses.  This further improves {\it utility} by not
  showing suggestions when they are unlikely to be used.  We break this out into
  a separate component so that we have the option to use a computationally
  cheaper architecture than what is used for the scoring model; this keeps
  the system {\it scalable}.  This model is described in Section
  \ref{sec:triggering}.

\end{enumerate}

The combination of these components is a novel end-to-end method for generating
short, complete responses to emails, going beyond previous works.  For response selection it exploits state-of-the-art deep learning
models trained on billions of words, and for response set generation it
introduces a new semi-supervised method for semantic understanding of
user-generated content.

Moreover, since it directly addresses all four challenges mentioned above, it
has been successfully deployed in {\emph Inbox}. Currently, the Smart Reply system is
responsible for assisting with $10$\% of email replies for {\emph Inbox} on mobile.

Next, we discuss the related work in Section~\ref{sec:related}, followed by a
description of our core system components in Sections
\ref{sec:response_selection}, \ref{sec:response_generation},
\ref{sec:diversity}, and \ref{sec:triggering}.  We close by showing modeling
results in Section~\ref{sec:eval_results} and our conclusions in
Section~\ref{sec:conc}.

\section{Related work}
\label{sec:related}

As we consider related work, we note that building an automated system to
suggest email responses is not a task for which there is existing literature
or benchmarks, nor is this a standard machine learning problem to which
existing algorithms can be readily applied.  However, there is work related to
two of our core components which we will review here: predicting responses and
identifying a target response space. \\

\noindent{\bf Predicting full responses.}  Much work exists on analyzing
natural language dialogues in public domains such as Twitter, but it has largely
focused on social media tasks like predicting whether or not a response is made
\cite{artzi-pantel-gamon:2012}, predicting next word only
\cite{pang-ravi:2012}, or curating threads \cite{Backstrom:2013}.

Full response prediction was initially attempted in \cite{ritter}, which
approached the problem from the perspective of machine translation: given a
Twitter post, "translate" it into a response using phrase-based statistical machine translation (SMT).  Our
approach is similar, but rather than using SMT we use the neural network machine
translation model proposed in \cite{SutskeverVL14}, called
"sequence-to-sequence learning".

Sequence-to-sequence learning, which makes use of long short-term memory
networks (LSTMs) \cite{lstm} to predict sequences of text, was originally
applied to Machine Translation but has since seen success in other domains such as image
captioning \cite{vinyals_image} and speech recognition \cite{chan}.

Other recent works have also applied recurrent neural networks (RNNs) or LSTMs
to full response prediction \cite{sordoni}, \cite{shang}, \cite{serban},
\cite{vinyals_conversation}.  In \cite{sordoni} the authors rely on having an SMT
system to generate n-best lists, while \cite{serban} and
\cite{vinyals_conversation}, like this work, develop fully generative models.
Our approach is most similar to the Neural Conversation Model
\cite{vinyals_conversation}, which uses sequence-to-sequence learning to model
tech support chats and movie subtitles.

The primary difference of our work is that it was deployed in a production
setting, which raised the challenges of response quality, utility, scalability,
and privacy.  These challenges were not considered in any of these related works
and led to our novel solutions explained in the rest of this paper.

Furthermore, in this work we approach a different domain than
\cite{sordoni}, \cite{shang}, \cite{serban}, and \cite{vinyals_conversation}, which
primarily focus on social media and movie dialogues.
In both of those domains it can be acceptable to provide a
response that is merely related or on-topic.  Email, on the other hand,
frequently expresses a request or intent which must be addressed in the
response. \\

\noindent{\bf Identifying a target response space.}  Our approach here builds on
the Expander graph learning approach \cite{expander}, since it scales well to
both large data (vast amounts of email) and large output sizes (many different
underlying semantic intents).  While Expander was originally proposed
for knowledge expansion and classification tasks \cite{emailcateg2016}, our work
is the first to use it to discover semantic intent clusters from user-generated
content.

Other graph-based semi-supervised learning techniques have been explored in the
past for more traditional classification problems
\cite{Zhu03ssl, Bengio+al-ssl-2006}.  Other related works have explored tasks
involving semantic classification~\cite{li:2010} or identifying word-level
intents~\cite{SahaRoy:2015} targeted towards Web search queries and other
forums~\cite{chen-EtAl:2013}. However, the problem settings and tasks
themselves are significantly different from what is addressed in our work. \\

Finally, we note that Smart Reply is the first work to address these tasks
together and solve them in a single end-to-end, deployable system.

\section{Selecting responses}
\label{sec:response_selection}

The fundamental task of the Smart Reply system is to find the most likely
response given an original message.  In other words, given original message
$\mathbf{o}$ and the set of all possible responses $R$, we would like to find:

\[\mathbf{r^*} =\underset{r \in R}{\mathrm{argmax}} \hspace{2pt} P(\mathbf{r}|\mathbf{o})\]

To find this response, we will construct a model that can score responses and
then find the highest scoring response.

We will next describe how the model is formulated, trained, and used for inference.
Then we will discuss the core challenges of bringing this model to produce high quality
suggestions on a large scale.

\subsection{LSTM model}

Since we are scoring one sequence of tokens $\mathbf{r}$,
conditional on another sequence of tokens $\mathbf{o}$, this problem is a
natural fit for sequence-to-sequence learning \cite{SutskeverVL14}.  The model
itself is an LSTM.  The input is the tokens of the original message
$\{o_1, ..., o_n\}$, and the output is the conditional probability distribution
of the sequence of response tokens given the input:
\vspace{-0.3cm}
\[P(r_1, ..., r_m | o_1, ..., o_n)\]

As in \cite{SutskeverVL14}, this distribution can be factorized as:

\[P(r_1, ..., r_m | o_1, ..., o_n) = \prod_{i=1}^mP(r_i|o_1, ..., o_n, r_1, ..., r_{i-1})  \]

First, the sequence of original message tokens, including a special
end-of-message token $o_n$, are read in, such that the LSTM's hidden state encodes a
vector representation of the whole message.  Then, given this hidden state, a
softmax output is computed and interpreted as $P(r_1|o_1, ..., o_n)$, or the
probability distribution for the first response token.  As response tokens are
fed in, the softmax at each timestep $t$ is interpreted as
$P(r_t|o_1, ..., o_n, r_1, ..., r_{t-1})$.  Given the factorization above,
these softmaxes can be used to compute $P(r_1, ..., r_m | o_1, ..., o_n)$. \\

\noindent{\bf Training}  Given a large corpus of messages, the training objective
is to maximize the log probability of observed responses, given their
respective originals:

\[ \sum_{(\mathbf{o}, \mathbf{r})}  \log P(r_1, ..., r_m | o_1, ..., o_n)\]

We train against this objective using stochastic gradient descent with AdaGrad
\cite{duchi}.  Ten epochs are run over a message corpus which will be described
in Section \ref{sec:data}.  Due to the size of the corpus, training is run in
a distributed fashion using the TensorFlow library \cite{tensorflow}.

Both our input and output vocabularies consist of the most frequent
English words in our training data after preprocessing (steps described in Section~\ref{sec:data}).  In addition to the standard LSTM
formulation, we found that the addition of a recurrent projection layer
\cite{sak} substantially improved both the quality of the converged model and
the time to converge.  We also found that gradient clipping (with the value of
$1$) was essential to stable training.  \\

\noindent{\bf Inference} At inference time we can feed in an original
message and then use the output of the softmaxes to get a probability
distribution over the vocabulary at each timestep.  These distributions can be
used in a variety of ways:

\begin{enumerate}

\item To draw a random sample from the response distribution $P(r_1, ..., r_m
  | o_1, ..., o_n)$.  This can be done by sampling one token at each timestep
  and feeding it back into the model.

\item To approximate the most likely response, given the original message.  This
  can be done greedily by taking the most likely token at each time step and
  feeding it back in.  A less greedy strategy is to use a beam search, i.e.,
  take the top $b$ tokens and feed them in, then retain the $b$ best response prefixes
  and repeat.

\item To determine the likelihood of a specific response candidate.  This can be
  done by feeding in each token of the candidate and using the softmax output to
  get the likelihood of the next candidate token.

\end{enumerate}

Table \ref{fig:generative_examples} shows some example of generating the
approximate most likely responses using a beam search. \\

\begin{table}
\begin{tabular}{|p{3.5cm} |p{4.0 cm} |}
\hline
Query & Top generated responses\\
\hline
Hi, I thought it would be & I can do Tuesday. \\
great for us to sit down  & I can do Wednesday. \\
and chat.  I am free      & How about Tuesday? \\
Tuesday  and Wenesday.    & I can do Tuesday! \\
Can you do either of      & I can do Tuesday. What \\
those days?               & time works for you? \\
                          & I can do Wednesday! \\
Thanks!                   & I can do Tuesday or  \\
                          & Wednesday. \\
--Alice                   & How about Wednesday? \\
                          & I can do Wednesday. What time works for you?\\
                          & I can do either.\\
\hline
\end{tabular}
\caption{Generated response examples.}
\label{fig:generative_examples}
\end{table}

\subsection{Challenges}\label{sec:lstm_challenges}

As described thus far, the model can generate coherent and plausible responses given an
incoming email message.  However, several key challenges arise when bringing
this model into production. \\

\noindent{\bf Response quality}  In order to surface responses to users, we need to
ensure that they are {\it always} high quality in style, tone, diction, and
content.

Given that the model is trained on a corpus of real messages, we
have to account for the possibility that the most probable response is not
necessarily a high quality response.  Even a response that occurs frequently in
our corpus may not be appropriate to surface back to users.  For example, it
could contain poor grammar, spelling, or mechanics ({\it your the best!}); it
could convey a familiarity that is likely to be jarring or offensive in many
situations ({\it thanks hon!}); it could be too informal to be consistent with
other {\it Inbox} intelligence features ({\it yup, got it thx}); it could convey
a sentiment that is politically incorrect, offensive, or otherwise
inappropriate ({\it Leave me alone}).

While restricting the model vocabulary might address simple cases such as
profanity and spelling errors, it would not be sufficient to capture the wide
variability with which, for example, politically incorrect statements can be
made.  Instead, we use semi-supervised learning (described in detail in Section
\ref{sec:response_generation}) to
construct a target response space $R$ comprising only high quality responses.
Then we use the model described here to choose the best response in $R$, rather
than the best response from any sequence of words in its vocabulary.  \\

\noindent{\bf Utility} Our user studies showed that suggestions are most useful
when they are highly specific to the original message and express diverse
intents.  However, as column 1 in Table ~\ref{fig:eg_ranked_responses} shows,
the raw output of the model tends to (1) favor common but unspecific responses and (2)
have little diversity.

First, to improve specificity of responses, we apply some light normalization
that penalizes responses which are applicable to a broad range of incoming
messages.  The results of this normalization can be seen in column 2 of
Table~\ref{fig:eg_ranked_responses}.  For example, the very generic "Yes!" has
fallen out of the top ten.  Second, to increase the breadth of options shown to
the user, we enforce diversity by exploiting the semantic structure of $R$, as
we will discuss in Section \ref{sec:diversity}.  The results of this are also shown at
the bottom of Table~\ref{fig:eg_ranked_responses}.

We further improve the utility of suggestions by first passing each message
through a triggering model  (described in Section \ref{sec:triggering}) that
determines whether suggestions should be generated at all.  This reduces the
likelihood that we show suggestions when they would not be used anyway.  \\

\noindent{\bf Scalability}  Our model needs to be deployed in a production
setting and cannot introduce latency to the process of email delivery, so
scalability is critical.

Exhaustively scoring every response candidate $r \in R$,
would require $O(|R|l)$ LSTM steps where $l$ is the length of the longest
response.  In previous work \cite{SutskeverVL14}, this could be afforded
because computations were performed in a batch process offline.  However,
in the context of an email delivery pipeline, time is a much more
precious resource.  Furthermore, given the tremendous diversity with which people
communicate and the large number of email scenarios we would like to cover,
we can assume that $R$ is very large and only expected to grow over time.
For example, in a uniform sample of 10 million short responses (say, responses
with at most 10 tokens), more than $40$\% occur only once.
Therefore, rather than performing an exhaustive scoring of every candidate
$r \in R$, we would like to efficiently search for the best responses
such that complexity is $not$ a function of $|R|$.

\begin{figure}
\begin{tikzpicture}
    \begin{axis}[
        xlabel=Beam size,
        ylabel=Frequency of matching exhaustive search]
    \addplot[smooth,mark=*,blue] plot coordinates {
    (1, 0.128)
    (2, 0.297)
    (4, 0.604)
    (8, 0.793)
    (16, 0.931)
    (32, 0.961)
    (64, 0.98)
    (128, 0.987)
    };
    \end{axis}
\end{tikzpicture}
\caption{Effectiveness of searching the response space $R$.  {\normalfont For a
         sample of messages we compute the frequency with which the best
         candidate found by a beam search over $R$ matches the best candidate
         found by exhaustively scoring all members of $R$.  We compare various
         beam sizes.  At a beam size of $16$, these two methods find the
         same best response $93\%$ of the time.}}
\label{fig:beam_search}
\end{figure}
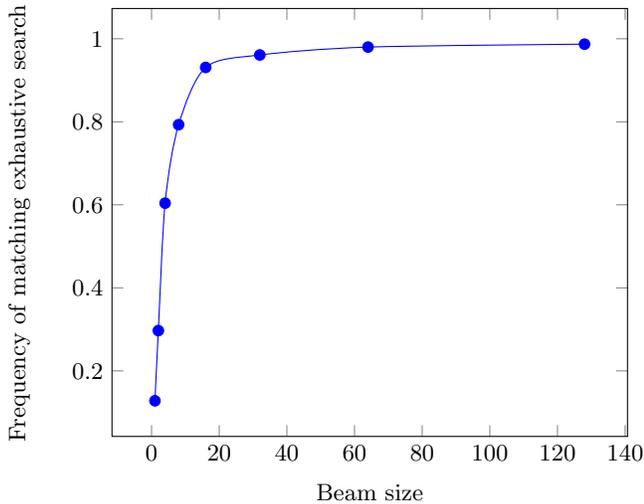

Our search is conducted as follows.  First, the elements of $R$
are organized into a trie.  Then, we conduct a left-to-right beam search,
but only retain hypotheses that appear in the trie.  This search process has
complexity $O(bl)$ for beam size $b$ and maximum response length $l$.
Both $b$ and $l$ are typically in the range of 10-30, so this method
dramatically reduces the time to find the top responses and is a critical
element of making this system deployable.  In terms of quality, we find that,
although this search only approximates the best responses in $R$, its results
are very similar to what we would get by scoring and ranking all $r \in R$,
even for small $b$.  At $b = 128$, for example, the top scoring
response found by this process matches the true top scoring response
$99\%$ of the time.  Results for various beam sizes are shown in Figure
\ref{fig:beam_search}.

Additionally, requiring that each message first pass through a triggering model,
as mentioned above, has the additional benefit of reducing the total amount
of LSTM computation. \\

\noindent{\bf Privacy} Note that all email data (raw data, preprocessed data and training data) was
encrypted. Engineers could only inspect aggregated statistics on anonymized
sentences that occurred across many users and did not identify any user.
Also, only frequent words are retained. As a result, verifying model's
quality and debugging is more complex.\\

Our solutions for the first three challenges are described further in Sections
\ref{sec:response_generation}, \ref{sec:diversity}, and
\ref{sec:triggering}.

\newpage
\section{Response Set Generation}
\label{sec:response_generation}

Two of the core challenges we face when building the end to end automated
response system are {\it response quality} and {\it utility}.  Response quality comes from
suggesting ``high quality'' responses that deliver a positive user experience.
Utility comes from ensuring that we don't suggest
multiple responses that capture the same intent (for example, minor lexical
variations such as ``{\it Yes, I'll be there.}'' and ``{\it I will be there.}'').
We can consider these two challenges jointly.

We first need to define a target response space that
comprises high quality messages which can be surfaced as suggestions. The goal
here is to generate a structured response set that effectively captures various
intents conveyed by people in natural language conversations. The target
response space should capture both variability in language and intents. The
result is used in two ways downstream---(a) define a response space for scoring
and selecting suggestions using the model described in
Section~\ref{sec:response_selection}, and (b) promote diversity among chosen
suggestions as discussed in Section~\ref{sec:diversity}.

We construct a response set using only the most frequent anonymized sentences
aggregated from the preprocessed data (described in
Section~\ref{sec:data}). This process yields a few million unique sentences.

\subsection{Canonicalizing email responses}

The first step is to automatically generate a set of canonical responses
messages that capture the variability in language. For example, responses such
as ``{\it Thanks for your kind update.}'', ``{\it Thank you for updating!}'',
``{\it Thanks for the status update.}'' may appear slightly different on the
surface but in fact convey the same information. We parse each sentence using
a dependency parser and use its syntactic structure to generate a canonicalized
representation. Words (or phrases) that are modifiers or unattached to head
words are ignored.

\subsection{Semantic intent clustering}

In the next step, we want to partition all response messages into
``semantic'' clusters where a cluster represents a meaningful response
intent (for example, ``{\it thank you}'' type of response versus ``{\it sorry}''
versus ``{\it cannot make it}''). All messages within a cluster share the same
semantic meaning but may appear very different. For example, ``{\it Ha ha}'',
``{\it lol}'' and ``{\it Oh that's funny!}'' are associated with the {\it
funny} cluster.

This step helps to automatically digest the entire information present in
frequent responses into a coherent set of semantic clusters. If we were
to build a semantic intent prediction model for this purpose, we would need
access to a large corpus of sentences annotated with their corresponding
semantic intents. However, this is neither readily available for our task nor at
this scale. Moreover, unlike typical machine learning classification tasks, the
semantic intent space cannot be fully defined a priori. So instead, we model the
task as a semi-supervised machine learning problem and use scalable graph
algorithms~\cite{expander} to automatically learn this information from data and
a few human-provided examples.

\subsection{Graph construction}
We start with a few manually defined clusters sampled from the top frequent
messages (e.g., {\it thanks}, {\it i love you}, {\it sounds good}). A small
number of example responses are added as ``seeds'' for each cluster (for
example, {\it thanks} $\rightarrow$ ``{\it Thanks!}'', ``{\it Thank
you.}'').\footnote{In practice, we pick $100$ clusters and on average $3$--$5$
labeled seed examples per cluster.}

\begin{figure*}[ht!]
\centering
  \includegraphics[width=0.85\textwidth,height=6.3cm]{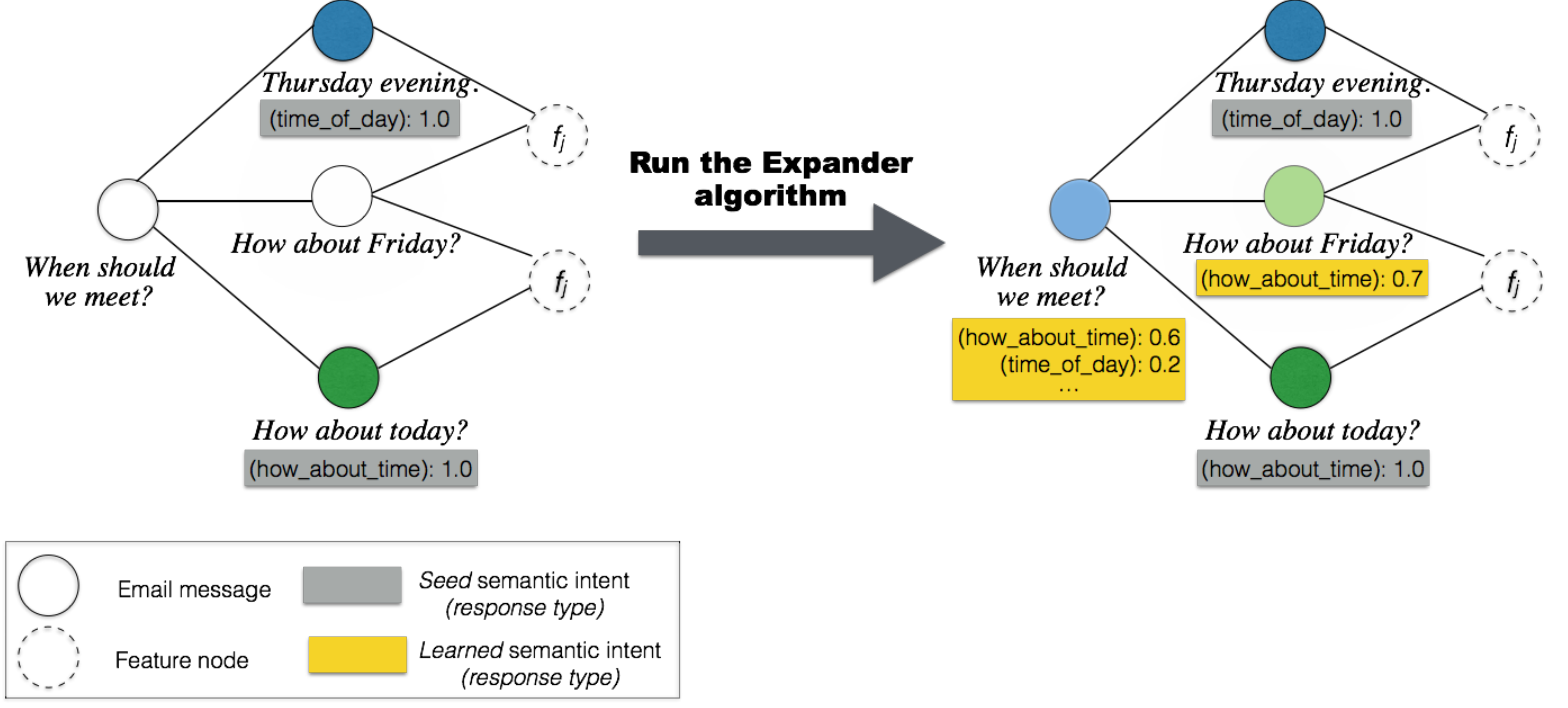}
  \caption{Semantic clustering of response messages.}
  \label{fig:clustering}
\end{figure*}

We then construct a base graph with frequent response messages as nodes ($V_R$).
For each response message, we further extract a set of lexical features (ngrams
and skip-grams of length up to 3) and add these as ``feature'' nodes ($V_F$) to
the same graph. Edges are created between a pair of nodes $(u, v)$ where $u \in
V_R$ and $v \in V_F$ if $v$ belongs to the feature set for response $u$. We
follow the same process and create nodes for the manually labeled examples
$V_L$. We make an observation that in some cases an incoming original message
could potentially be treated as a response to another email depending on the
context. For example, consider the following (original, response) message pairs:
\begin{center}
\noindent{\it Let us get together soon.} $\rightarrow$ {\it When should we meet?}\\
\noindent{\it When should we meet?} $\rightarrow$ {\it How about Friday?}\\
\end{center}
Inter-message relations as shown in the above example can be modeled within the same framework by adding extra edges between the corresponding message nodes in the graph.

\subsection{Semi-supervised learning}
The constructed graph captures
relationships between similar canonicalized responses via the feature nodes.
Next, we learn a semantic labeling for all response nodes by propagating
semantic intent information from the manually labeled examples through the
graph. We treat this as a semi-supervised learning problem and use the
distributed EXPANDER~\cite{expander} framework for optimization. The learning
framework is scalable and naturally suited for semi-supervised graph propagation
tasks such as the semantic clustering problem described here. We minimize the
following objective function for response nodes in the graph:

\begin{equation}
\begin{split}
  &s_{i}||\hat{C}_i - C_i||^2 + \mu_{pp}||\hat{C}_i - U||^2 \\
  &+ \mu_{np} \Big( \sum_{j\in {\cal N_F}(i)} w_{ij}||\hat{C}_i - \hat{C}_j||^2 + \sum_{j\in {\cal N_R}(i)}w_{ik}||\hat{C}_i - \hat{C}_k||^2 \Big)
\end{split}
\end{equation}
where $s_{i}$ is an indicator function equal to 1 if the node $i$ is a seed and 0 otherwise,
$\hat{C}_i$ is the learned semantic cluster distribution for response node $i$,
$C_i$ is the true label distribution (i.e., for manually provided examples),
${\cal N_F}(i)$ and ${\cal N_R}(i)$ represent the feature and message neighborhood of node $i$,
$\mu_{np}$ is the predefined penalty for neighboring nodes with divergent label distributions,
$\hat{C}_j$ is the learned label distribution for feature neighbor $j$,
$w_{ij}$ is the weight of feature $j$ in response $i$,
$\mu_{pp}$ is the penalty for label distribution deviating from the prior, a uniform distribution $U$.

The objective function for a feature node is alike, except that there is no first term, as there are no seed labels for feature nodes:
\begin{equation}
\mu_{np}\sum_{i\in {\cal N}(j)}w_{ij}||\hat{C}_j - \hat{C}_i||^2 + \mu_{pp}||\hat{C}_j - U||^2
\end{equation}

The objective function is jointly optimized for all nodes in the graph.

The output from EXPANDER is a learned distribution of semantic labels for every node in the graph. We assign the top scoring output label as the semantic intent for the node, labels with low scores are filtered out. Figure~\ref{fig:clustering} illustrates this process.

To discover new clusters which are not covered by the labeled examples, we run
the semi-supervised learning algorithm in repeated phases. In the first phase,
we run the label propagation algorithm for 5 iterations. We then fix the cluster
assignment, randomly sample 100 new responses from the remaining unlabeled nodes
in the graph. The sampled nodes are treated as potential new clusters and
labeled with their canonicalized representation. We rerun label propagation with
the new labeled set of clusters and repeat this procedure until convergence
(i.e., until no new clusters are discovered and members of a cluster do not
change between iterations). The iterative propagation method thereby allows us
to both expand cluster membership as well as discover (up to 5X) new clusters,
where each cluster has an interpretable semantic interpretation.

\subsection{Cluster Validation}
Finally, we extract the top $k$ members for each semantic cluster, sorted by
their label scores. The set of (response, cluster label) pairs are then
validated by human raters. The raters are provided with a response $R_i$,
a corresponding cluster label $C$ (e.g., {\it thanks}) as well as few example
responses belonging to the cluster (e.g., ``{\it Thanks!}'', ``{\it Thank
you.}'') and asked whether $R_i$ belongs to $C$.

The result is an automatically generated and validated set of high quality
response messages labeled with semantic intent. This is subsequently used by the
response scoring model to search for approximate best responses to an incoming
email (described earlier in Section~\ref{sec:response_selection}) and further to enforce diversity among the top
responses chosen (Section~\ref{sec:diversity}).

\section{Suggestion Diversity}
\label{sec:diversity}

As discussed in Section~\ref{sec:response_selection}, the LSTM first processes an incoming
message and then selects the (approximate) best responses from the target response
set created using the method described in Section~\ref{sec:response_generation}.
Recall that we follow this by some light normalization to penalize responses
that may be too general to be valuable to the user.  The effect of this
normalization can be seen by comparing columns 1 and 2 of
Table~\ref{fig:eg_ranked_responses}.  For example, the very generic "Yes!" falls
out of the top ten responses.

Next, we need to choose a small number of options to show the user.
A straight-forward approach would be to just choose
the $N$ top responses and present them to the user. However, as column 2 of
Table~\ref{fig:eg_ranked_responses} shows, these responses tend to be
very similar.

The likelihood of at least one response being useful is greatest when the
response options are not redundant, so it would be wasteful
to present the user with three variations of, say, \sentence{I'll be there.}
The job of the diversity component is to select a more varied set of
suggestions using two strategies: omitting redundant responses and
enforcing negative or positive responses.

\subsection{Omitting Redundant Responses}
This strategy assumes that the user should never see two responses of the
same \emph{intent}. An intent can be thought of as a cluster of responses that
have a common communication purpose, e.g. confirming, asking for time or
rejecting participation. In Smart Reply, every target response is
associated with exactly one intent. Intents are defined based on automatic
clustering followed by human validation as discussed in Section
\ref{sec:response_generation}.

The actual diversity strategy is simple: the top responses are iterated over in
the order of decreasing score. Each response is added to the list of
suggestions, unless its intent is already covered by a response on
the suggestion list. The resulting list contains only the highest-scored
representative of each intent, and these representatives are ordered by
decreasing score. \\

\subsection{Enforcing Negatives and Positives}
We have observed that the LSTM has a strong tendency towards producing positive
responses, whereas negative responses such as \sentence{I can't make it} or
\sentence{I don't think so} typically receive low scores. We believe that this
tendency reflects the style of email conversations: positive replies may be more
common, and where negative responses are appropriate, users may prefer a less
direct wording.

Nevertheless, we think that it is important to offer negative suggestions in
order to give the user a real choice. This policy is implemented through the
following strategy:

\begin{quote}
If the top two responses (after omitting redundant responses) contain at least
one positive response and none of the top three responses are negative, the
third response is replaced with a negative one.
\end{quote}

A positive response is one which is clearly affirmative, e.g. one that starts
with \emph{Yes}, \emph{Sure} or \emph{Of course}. In order to find the negative
response to be included as the third suggestion, a second LSTM pass is
performed. In this second pass, the search is restricted to only the negative
responses in the target set (refer Table~\ref{fig:eg_ranked_responses} for scored negative response examples). This is necessary since the top responses produced
in the first pass may not contain any negatives.

Even though missing negatives are more common, there are also cases in which
an incoming message triggers exclusively negative responses. In this situation,
we employ an analogous strategy for enforcing a positive response. \\

The final set of top scoring responses (bottom row in Table~\ref{fig:eg_ranked_responses}) are then presented to the user as suggestions.

\begin{table}
\begin{tabular}{|l||l|}
\hline
Unnormalized Responses & Normalized Responses\\
\hline
Yes, I'll be there. & Sure, I'll be there.\\
Yes, I will be there. & Yes, I can.\\
I'll be there. & Yes, I can be there.\\
Yes, I can. & Yes, I'll be there.\\
What time? & Sure, I can be there.\\
I'll be there! & Yeah, I can.\\
I will be there. & Yeah, I'll be there.\\
Sure, I'll be there. & Sure, I can.\\
Yes, I can be there. & Yes. I can.\\
Yes! & Yes, I will be there.\\
\hline
\hline
\multicolumn{2}{|l|}{Normalized Negative Responses} \\
\hline
\multicolumn{2}{|l|}{Sorry, I won't be able to make it tomorrow.} \\
\multicolumn{2}{|l|}{Unfortunately I can't.} \\
\multicolumn{2}{|l|}{Sorry, I won't be able to join you.} \\
\multicolumn{2}{|l|}{Sorry, I can't make it tomorrow.} \\
\multicolumn{2}{|l|}{No, I can't.} \\
\multicolumn{2}{|l|}{Sorry, I won't be able to make it today.} \\
\multicolumn{2}{|l|}{Sorry, I can't.} \\
\multicolumn{2}{|l|}{I will not be available tomorrow.} \\
\multicolumn{2}{|l|}{I won't be available tomorrow.} \\
\multicolumn{2}{|l|}{Unfortunately, I can't.} \\
\hline
\hline
\multicolumn{2}{|l|}{Final Suggestions} \\
\hline
\multicolumn{2}{|l|}{Sure, I'll be there.} \\
\multicolumn{2}{|l|}{Yes, I can.} \\
\multicolumn{2}{|l|}{Sorry, I won't be able to make it tomorrow.} \\
\hline
\end{tabular}
\caption{Different response rankings for the message
\emph{``Can you join tomorrow's meeting?''}
\label{fig:eg_ranked_responses}
}
\end{table}

\section{Triggering}
\label{sec:triggering}

The \emph{triggering} module is the entry point of the Smart Reply system.
It is responsible for filtering messages that are bad candidates for suggesting
responses. This includes emails for which short replies are not appropriate (e.g.,
containing open-ended questions or sensitive topics), as well as emails for
which no reply is necessary at all (e.g., promotional emails and auto-generated
updates).

The module is applied to every incoming email just after the
preprocessing step. If the decision is negative, we finish the execution and do
not show any suggestions (see Figure~\ref{fig:inference_overview}). Currently,
the system decides to produce a Smart Reply for roughly $11$\% of messages, so
this process vastly reduces the number of useless sugestions seen by the users.
An additional benefit is to decrease the number of calls to the more expensive
LSTM inference, which translates into smaller infrastructure cost.

There are two main requirements for the design of the triggering component.
First, it has to be good enough to figure out cases where the response is not
expected. Note that this is a very different goal than just scoring a set of
responses. For instance, we could propose several valid replies to a newsletter
containing a sentence ``{\it Where do you want to go today?}'', but most likely
all of the responses would be useless for our users.  Second, it has to be
fast: it processes hundreds of millions of messages daily, so we aim to
process each message within milliseconds.

The main part of the triggering component is a feedforward neural network which
produces a probability score for every incoming message. If the score is above
some threshold, we {\it trigger} and run the LSTM scoring.
We have adopted this approach because feedforward networks have repeatedly been shown
to outperform linear models such as SVM or linear regression on various NLP tasks (see
for example \cite{DBLP:journals/corr/Goldberg15c}).

\subsection{Data and Features}\label{triggering:data_features}
In order to label our training corpus of emails, we use as positive examples
those emails that have been responded to. More precisely, out of the data set described
in Section~\ref{sec:data}, we create a training set that consists of pairs
($\mathbf{o}$, $y$), where $\mathbf{o}$ is an incoming message and $y \in
\{true, false\}$ is a boolean label, which is {\it true} if the message had a response and {\it false} otherwise.  For the
positive class, we consider only messages that were replied to from a mobile
device, while for negative we use a subset of all messages. We downsample
the negative class to balance the training set.  Our goal is to model
$P(y = true \mid \mathbf{o})$, the probability that message $\mathbf{o}$ will have a response on
mobile.

After preprocessing (described in Section~\ref{sec:data}), we extract content features (e.g. unigrams, bigrams) from the message body, subject and headers.  We also use
various social signals like whether the sender is in recipient's address
book, whether the sender is in recipient's social network and whether the
recipient responded in the past to this sender.

\subsection{Network Architecture and Training}

We use a feedforward multilayer perceptron with an embedding layer (for
a vocabulary of roughly one million words) and three fully connected hidden
layers. We use feature hashing to bucket rare words that are not present in
the vocabulary. The embeddings are separate for each sparse feature type (eg.
unigram, bigram) and within one feature type, we aggregate embeddings by summing
them up. Then, all sparse feature embeddings are concatenated with each other
and with the vector of dense features (those are real numbers and boolean values
mentioned in Section~\ref{triggering:data_features}).

We use the ReLu \cite{relu} activation function for non-linearity between layers.
The dropout \cite{dropout} layer is applied after each hidden layer. We train the model
using AdaGrad \cite{duchi} optimization algorithm with logistic loss cost
function. Similarly to the LSTM, the training is run in a distributed fashion
using the TensorFlow library \cite{tensorflow}.

\section{Evaluation and Results}
\label{sec:eval_results}

In this section, we describe the training and test data, as well as preprocessing steps used
for all messages. Then, we evaluate different components of the Smart Reply
system and present overall usage statistics.

\subsection{Data}
\label{sec:data}

To generate the training data for all Smart Reply models from sampled accounts, we
extracted all pairs of an incoming message and the user's response to that
message. For training the triggering model (see Section~\ref{sec:triggering}),
we additionally sampled a number of incoming personal messages which the user
didn't reply to. At the beginning of Smart Reply pipeline (Figure~\ref{fig:inference_overview}),
data is preprocessed in the following way:

\begin{description}
\itemsep-0.1em
\item[Language detection] The language of the message is identified and
non-English messages are discarded.
\item[Tokenization] Subject and message body are broken into words and
punctuation marks.
\item[Sentence segmentation] Sentences boundaries are identified in the
message body.
\item[Normalization] Infrequent words and entities like personal names, URLs,
email addresses, phone numbers etc. are replaced by special tokens.
\item[Quotation removal] Quoted original messages and forwarded messages are
removed.
\item[Salutation/close removal] Salutations like \sentence{Hi John} and closes
such as \sentence{Best regards, Mary} are removed.
\end{description}

After the preprocessing steps, the size of the training set is 238 million messages,
which include 153 million messages that have no response.

\subsection{Results} \label{sec:results}
The most important end-to-end metric for our system is the fraction of messages
for which it was used. This is currently $10$\% of all mobile replies.
Below we describe in more detail evaluation stats for different components of
the system. We evaluate all parts in isolation using both offline analysis as
well as online experiments run on a subset of accounts.

\subsubsection{Triggering results}

In order to evaluate the triggering model, we split the data set described in
Section~\ref{triggering:data_features} into train ($80$\%) and test ($20$\%)
such that all test messages are delivered after train messages. This is to
ensure that the test conditions are similar to the final scenario.
We use a set of standard binary classifier metrics: precision, recall and the area
under the ROC curve. The AUC of the triggering model is $0.854$.
We also compute the fraction of triggered messages in the deployed system, which is
$11$\%. We observed that it may be beneficial to slightly
over-trigger, since the cost of presenting a suggestion, even if it is not
used, is quite low.

\subsubsection{Response selection results} \label{sec:results_lstm}
We evaluate the LSTM scoring model on three standard metrics: Perplexity, Mean
Reciprocal Rank and Precision@K.

\paragraph{Perplexity}
Perplexity is a measure of how well the model has fit the data: a model with
lower perplexity assigns higher likelihood to the test responses, so we expect
it to be better at predicting responses. Intuitively, a perplexity equal to $k$
means that when the model predicts the next word, there are on average $k$
likely candidates. In particular, for the ideal scenario of perplexity equal to
$1$, we always know exactly what should be the next word.
The perplexity on a set of $N$ test samples is computed using the following formula:

\[ P_r =  \exp(- \frac{1}{W} \sum_{i=1}^{N} \ln(\hat{P}(r^{i}_{1}, ..., r^{i}_{m} | o^{i}_{1}, ..., o^{i}_{n}))) \]

\noindent where $W$ is the total number of words in all $N$ samples, $\hat{P}$ is the
learned distribution and $\mathbf{r}^i$, $\mathbf{o}^i$ are the $i$-th response
and original message. Note that in the equation above only response terms are factored into $P_r$.
The perplexity of the Smart Reply LSTM is $17.0$. By comparison, an
n-grams language model with Katz backoff \cite{katz} and a maximum order of $5$ has
a perplexity of $31.4$ on the same data (again, computed only from response terms).

\paragraph{Response ranking}

\newcommand{\argmin}{\operatornamewithlimits{argmin}}

While perplexity is a quality indicator, it does not actually measure
performance at the scoring task we are ultimately interested in.  In particular,
it does not take into account the constraint of choosing a response in $R$.
Therefore we also evaluate the model on a response ranking task: for each of
$N$ test message pairs $(o, r)$ for which $r \in R$, we
compute $s = P(r | o)$ and $\forall_i \: x_i = P(w_i | o)$, where $w_i$ is the
$i$-th element of $R$. Then we sort the set $R = \{s, x_1, \ldots, x_N \}$
in descending order. Finally, we define $rank_i = \argmin_j (R_j | R_j = s)$.
Put simply, we are finding the rank of the actual response with respect to
all elements in $R$.
\newpage
Using this value, we can compute the Mean Reciprocal Rank:
\vspace{-0.3cm}
\[ MRR = \frac{1}{N} \sum_{i=1}^{N} \frac{1}{rank_i} \]

Additionally we can compute Precision@K.
For a given value of $K$ it is computed as the number of cases for which target
response $r$ was within the top$K$ responses that were ranked by the model.

We compare the Smart Reply response selection model to three baselines on the
same ranking task.  The {\it Random} baseline ranks $R$ randomly.  The
 {\it Frequency} baseline ranks them in order of their frequency in the
training corpus.  This baseline captures the extent to which we can simply
suggest highly frequent responses without regard for the contents of the
original message.  The {\it Multiclass-BOW} baseline ranks $R$ using a
feedforward neural network whose input is the original message, represented
with bag of words features, and whose output is a distribution over the
elements of {$R$} (a softmax).

As shown in Table~\ref{fig:response_ranking}, the Smart Reply LSTM significantly
improves on the {\it Frequency} baseline, demonstrating that conditioning
on the original message is effective; the model successfully extracts information
from the original message and uses it to rank responses more accurately.

It also significantly outperforms the {\it Multiclass-BOW} baseline. There are
a few possible explanations for this.  First, the recurrent architecture allows the model to learn more
sophisticated language understanding than bag of words features.  Second, when we pose this as a mulitclass prediction
problem, we can only train on messages whose response is in $R$, a small fraction
of our data.  On the other hand, the sequence-to-sequence framework allows us to
take advantage of all data in our corpus: the model can learn a lot about
original-response relationships even when the response does not appear in $R$
exactly.

Note that an added disadvantage of the multiclass formulation is that it
tightly couples the training of the model to the construction of $R$.  We
expect $R$ to grow over time, given the incredible diversity with which
people communicate.  While a simpler application such as chat might only
need a small number of possible responses, we find that for email we will need
a tremendous number of possible suggestions to really address users' needs.

\begin{table}
\centering
\begin{tabular}{|c|c|c|c|}
  \hline
  Model & Precision@10 & Precision@20 & MRR \\
  \hline
  Random & $5.58e-4$ & $1.12e-3$ & $3.64e-4$ \\
  \hline
  Frequency & $0.321$ & $0.368$ & $0.155$ \\
  \hline
  Multiclass-BOW & $0.345$ & $0.425$ & $0.197$ \\
  \hline
  Smart Reply & $0.483$ & $0.579$ & $0.267$ \\
  \hline
\end{tabular}
\caption{Response ranking}\label{fig:response_ranking} 
\end{table}

\begin{figure}[t]
\centering
  \includegraphics[scale=0.7]{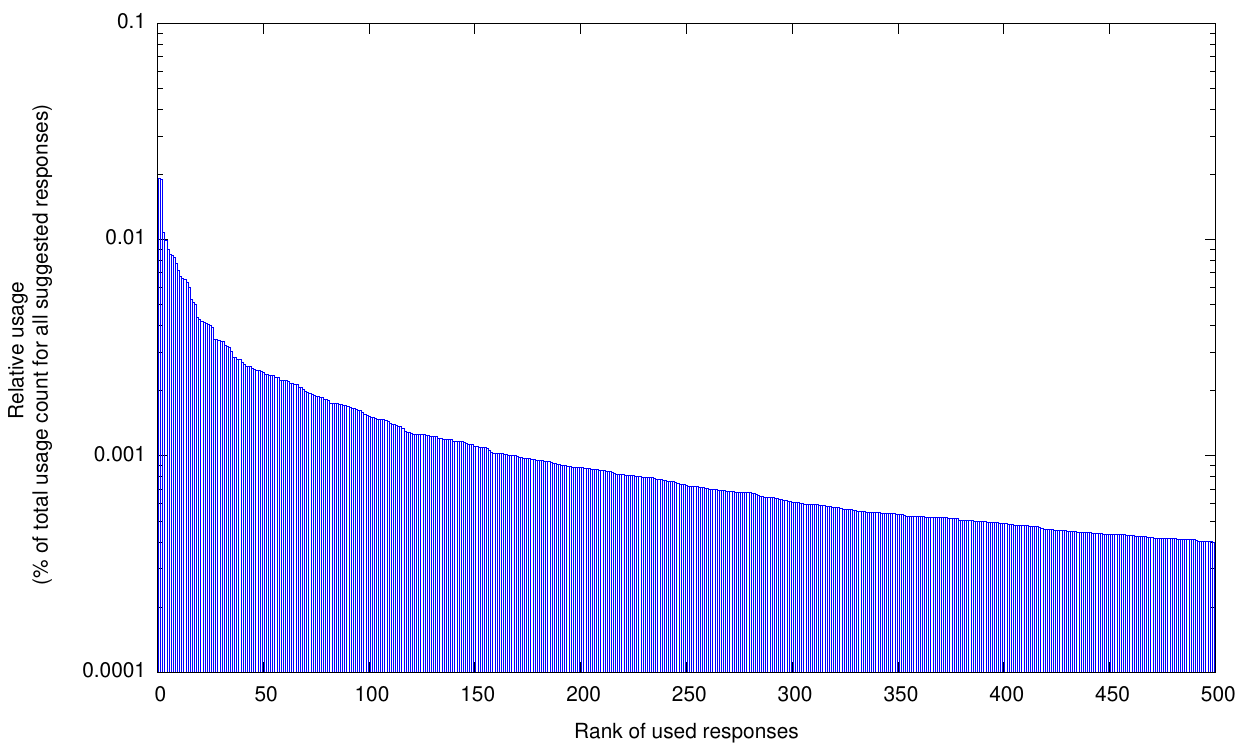}
  \caption{Usage distribution for top suggested responses.}
  \label{fig:used_responses}
\end{figure}

\subsubsection{Diversity results}

We justify the need for both the diversity component and a sizable response
space $R$ by reporting statistics around unique suggestions and clusters in
Table~\ref{fig:diversity_counts}.  The Smart Reply system generates daily
$12.9$k unique suggestions that belong to $376$ unique semantic clusters.
Out of those, people decide to use $4,115$, or $31.9$\% of, unique suggestions
and $313$, or $83.2$\% of, unique clusters. Note, however, that many suggestions are never
seen, as column 2 shows: the user may not open an email, use the web interface
instead of mobile or just not scroll down to the bottom of the message. Also, only
one of the three displayed suggestions will be selected by the user. These
statistics demonstrate the need to go well beyond a simple system with 5 or 10
canned responses.

\begin{table}
\centering
\begin{tabular}{|c|c|c|c|}
  \hline
     & Daily Count & Seen & Used \\
  \hline
  Unique Clusters & $376$ & $97.1$\% & $83.2$\% \\
  \hline
  Unique Suggestions & $12.9$k & $78$\% & $31.9$\% \\
  \hline
\end{tabular}
\caption{Unique cluster/suggestions usage per day}\label{fig:diversity_counts}
\end{table}

\begin{figure}[t]
\centering
  \includegraphics[scale=0.7]{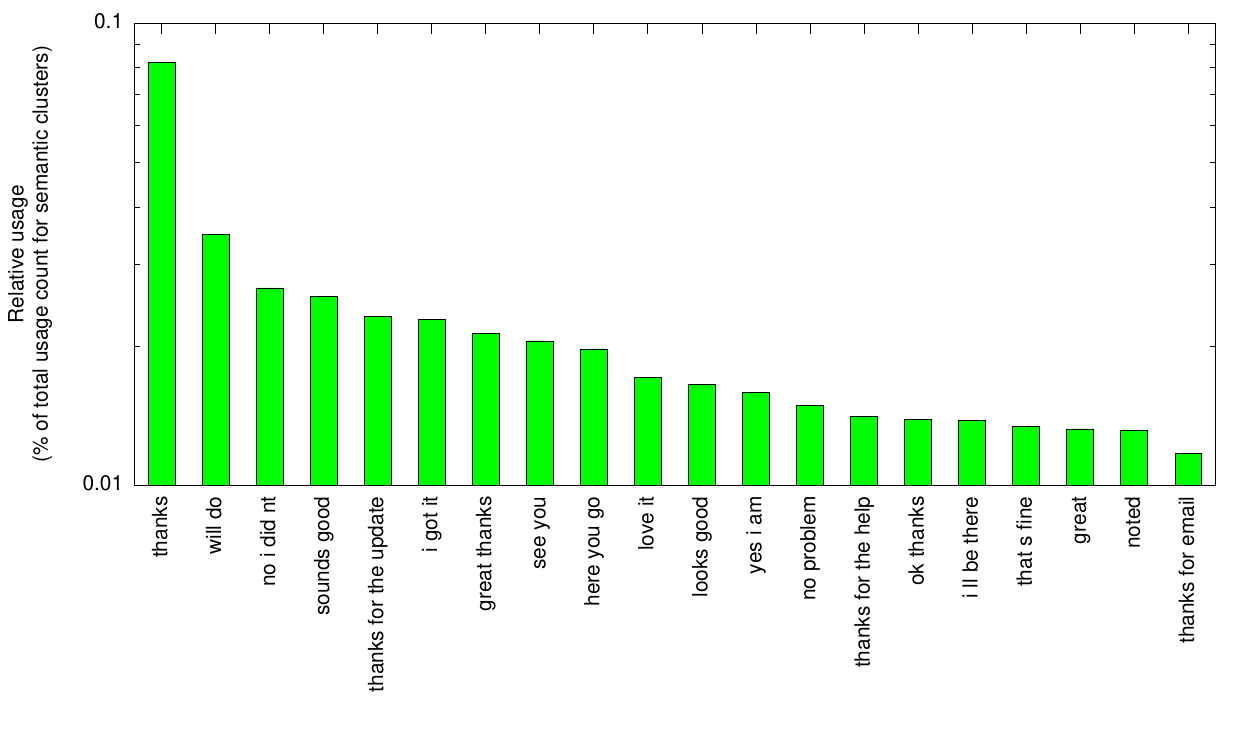}
  \caption{Usage distribution for semantic clusters corresponding to top suggested responses.}
  \label{fig:used_clusters}
\end{figure}

Figure~\ref{fig:used_responses} and Figure~\ref{fig:used_clusters} present,
respectively, the distribution of the rank for suggested responses and the
distribution of suggested clusters. The tail of the cluster distribution is
long, which explains the poor performance of {\it Frequency} baseline described in
Section~\ref{sec:results_lstm}.

We also measured how Smart Reply suggestions are used based on their
location on a screen. Recall that Smart Reply always presents $3$ suggestions,
where the first suggestion is the top one. We observed that, out of all used
suggestions, $45$\% were from the $1st$ position, $35$\% from the $2$nd position
and $20$\% from the $3$rd position. Since usually the third position is used
for diverse responses, we conclude that the diversity component is
crucial for the system quality.

Finally, we measured the impact of enforcing a diverse set of responses (e.g.,
by not showing two responses from the same semantic cluster) on user engagement:
when we completely disabled the diversity component and simply suggested the
three suggestions with the highest scores, the click-through rate decreased by
roughly $7.5$\% relative.

\section{Conclusions}
\label{sec:conc}

We presented Smart Reply, a novel end-to-end system for automatically
generating short, complete email responses.  The core of the system is a
state-of-the-art deep LSTM model that can predict full responses, given an
incoming email message.   To successfully deploy this system in {\it Inbox by
Gmail}, we addressed several challenges:
\begin{itemize}
\item{We ensure that individual response options deliver {\it quality} by
selecting them from a carefully constructed response space. The responses are
identified by a novel method for semantic clustering.}
\item{We increase the total {\it utility} of our chosen combination of suggestions
by enforcing diversity among them, and filtering traffic for which suggestions
would not be useful.}
\item{We build a {\it scalable} system by efficiently searching the target
response space.}
\end{itemize}

Our clearest metric of success is the fact that $10\%$ of mobile replies in
{\it Inbox} are now composed with assistance from the Smart Reply system.
Furthermore, we have designed the system in such a way that it is easily
extendable to address additional user needs; for instance, the
architecture of our core response scoring model is language agnostic,
therefore accommodates extension to other languages in addition to English.

\section{Acknowledgments}
The authors would like to thank Oriol Vinyals and Ilya Sutskever for
many helpful discussions and insights, as well as Prabhakar Raghavan
for his guidance and support.

\bibliographystyle{abbrv}
\scriptsize
\bibliography{smartreply}  

\begin{thebibliography}{10}

\bibitem{tensorflow}
M.~Abadi, A.~Agarwal, P.~Barham, and et~al.
\newblock Tensorflow: Large-scale machine learning on heterogeneous systems.
\newblock 2015.

\bibitem{ipsos}
I.~G.~P. Affairs.
\newblock Interconnected world: Communication \& social networking.
\newblock Press Release, March 2012.
\newblock \url{http://www.ipsos-na.com/news-polls/pressrelease.aspx?id=5564}.

\bibitem{artzi-pantel-gamon:2012}
Y.~Artzi, P.~Pantel, and M.~Gamon.
\newblock Predicting responses to microblog posts.
\newblock In {\em Proceedings of the 2012 Conference of the North American
  Chapter of the Association for Computational Linguistics: Human Language
  Technologies}, pages 602--606, Montr\'{e}al, Canada, June 2012. Association
  for Computational Linguistics.

\bibitem{Backstrom:2013}
L.~Backstrom, J.~Kleinberg, L.~Lee, and C.~Danescu-Niculescu-Mizil.
\newblock Characterizing and curating conversation threads: Expansion, focus,
  volume, re-entry.
\newblock In {\em Proceedings of the Sixth ACM International Conference on Web
  Search and Data Mining}, WSDM '13, pages 13--22, 2013.

\bibitem{Bengio+al-ssl-2006}
Y.~Bengio, O.~Delalleau, and N.~{Le Roux}.
\newblock Label propagation and quadratic criterion.
\newblock In O.~Chapelle, B.~Sch{\"o}lkopf, and A.~Zien, editors, {\em
  Semi-Supervised Learning}, pages 193--216. {MIT} Press, 2006.

\bibitem{chan}
W.~Chan, N.~Jaitly, Q.~V. Le, and O.~Vinyals.
\newblock Listen, attend, and spell.
\newblock {\em arXiv:1508.01211}, abs/1508.01211, 2015.

\bibitem{chen-EtAl:2013}
Z.~Chen, B.~Liu, M.~Hsu, M.~Castellanos, and R.~Ghosh.
\newblock Identifying intention posts in discussion forums.
\newblock In {\em Proceedings of the 2013 Conference of the North American
  Chapter of the Association for Computational Linguistics: Human Language
  Technologies}, pages 1041--1050, Atlanta, Georgia, June 2013. Association for
  Computational Linguistics.

\bibitem{duchi}
J.~Duchi, E.~Hazad, and Y.~Singer.
\newblock Adaptive subgradient methods for online learning and stochastic
  optimization.
\newblock {\em JMLR}, 12, 2011.

\bibitem{DBLP:journals/corr/Goldberg15c}
Y.~Goldberg.
\newblock A primer on neural network models for natural language processing.
\newblock {\em CoRR}, abs/1510.00726, 2015.

\bibitem{lstm}
S.~Hochreiter and J.~Schmidhuber.
\newblock Long short-term memory.
\newblock {\em Neural Computation}, 9(8):1735--1780, 1997.

\bibitem{katz}
S.~M. Katz.
\newblock Estimation of probabilities from sparse data for the language model
  component of a speech recogniser.
\newblock {\em IEEE Transactions on Acoustics, Speech, and Signal Processing},
  35:400--401, 1987.

\bibitem{li:2010}
X.~Li.
\newblock Understanding the semantic structure of noun phrase queries.
\newblock In {\em Proceedings of the 48th Annual Meeting of the Association for
  Computational Linguistics (ACL)}, pages 1337--1345, Uppsala, Sweden, July
  2010. Association for Computational Linguistics.

\bibitem{relu}
V.~Nair and G.~E. Hinton.
\newblock Rectified linear units improve restricted boltzmann machines.
\newblock In {\em Proceedings of the 27th International Conference on Machine
  Learning (ICML-10)}, pages 807--814, 2010.

\bibitem{pang-ravi:2012}
B.~Pang and S.~Ravi.
\newblock Revisiting the predictability of language: Response completion in
  social media.
\newblock In {\em Proceedings of the 2012 Joint Conference on Empirical Methods
  in Natural Language Processing and Computational Natural Language Learning},
  pages 1489--1499, Jeju Island, Korea, July 2012. Association for
  Computational Linguistics.

\bibitem{expander}
S.~Ravi and Q.~Diao.
\newblock Large scale distributed semi-supervised learning using streaming
  approximation.
\newblock In {\em Proceedings of the 19th International Conference on
  Artificial Intelligence and Statistics (AISTATS)}, 2016.

\bibitem{ritter}
A.~Ritter, C.~Cherry, and W.~B. Dolan.
\newblock Data-driven response generation in social media.
\newblock In {\em Proceedings of the 2011 Conference on Empirical Methods in
  Natural Language Processing}, Edinburgh, UK, July 2011. Association for
  Computational Linguistics.

\bibitem{SahaRoy:2015}
R.~Saha~Roy, R.~Katare, N.~Ganguly, S.~Laxman, and M.~Choudhury.
\newblock Discovering and understanding word level user intent in web search
  queries.
\newblock {\em Web Semant.}, 30(C):22--38, Jan. 2015.

\bibitem{sak}
H.~Sak, A.~Senior, and F.~Beaufays.
\newblock Long short-term memory recurrent neural network architectures for
  large scale acoustic modeling.
\newblock In {\em Proceedings of the Annual Conference of International Speech
  Communication Association (INTERSPEECH)}, 2014.

\bibitem{serban}
I.~V. Serban, A.~Sordoni, Y.~Bengio, A.~Courville, and J.~Pineau.
\newblock Hierarchical neural network generative models for movie dialogues.
\newblock In {\em arXiv preprint arXiv:1507.04808}, 2015.

\bibitem{shang}
L.~Shang, Z.~Lu, and H.~Li.
\newblock Neural responding machine for short-text conversation.
\newblock In {\em In Proceedings of ACL-IJCNLP}, 2015.

\bibitem{sordoni}
A.~Sordoni, M.~Galley, M.~Auli, C.~Brockett, Y.~Ji, M.~Mitchell, J.-Y. Nie,
  J.~Gao, and B.~Dolan.
\newblock A neural network approach to context-sensitive generation of
  conversation responses.
\newblock In {\em In Proceedings of NAACL-HLT}, 2015.

\bibitem{dropout}
N.~Srivastava, G.~Hinton, A.~Krizhevsky, I.~Sutskever, and R.~Salakhutdinov.
\newblock Dropout: A simple way to prevent neural networks from overfitting.
\newblock {\em Journal of Machine Learning Research}, 15:1929--1958, 2014.

\bibitem{SutskeverVL14}
I.~Sutskever, O.~Vinyals, and Q.~V. Le.
\newblock Sequence to sequence learning with neural networks.
\newblock {\em Advances in Neural Information Processing Systems (NIPS)}, 2014.

\bibitem{vinyals_conversation}
O.~Vinyals and Q.~V. Le.
\newblock A neural conversation model.
\newblock In {\em ICML Deep Learning Workshop}, 2015.

\bibitem{vinyals_image}
O.~Vinyals, A.~Toshev, S.~Bengio, and D.~Erhan.
\newblock Show and tell: A neural image caption generator.
\newblock In {\em Proceedings of IEEE Conference on Computer Vision and Pattern
  Recognition (CVPR)}, 2015.

\bibitem{emailcateg2016}
J.~B. Wendt, M.~Bendersky, L.~Garcia-Pueyo, V.~Josifovski, B.~Miklos, I.~Krka,
  A.~Saikia, J.~Yang, M.-A. Cartright, and S.~Ravi.
\newblock Hierarchical label propagation and discovery for machine generated
  email.
\newblock In {\em Proceedings of the International Conference on Web Search and
  Data Mining (WSDM) (2016)}, 2016.

\bibitem{Zhu03ssl}
X.~Zhu, Z.~Ghahramani, and J.~Lafferty.
\newblock Semi-supervised learning using gaussian fields and harmonic
  functions.
\newblock In {\em Proceedings of the International Conference on Machine
  Learning (ICML)}, pages 912--919, 2003.

\end{thebibliography}
\end{document}